# Concurrent Linguistic Error Detection (CLED): a New Methodology for Error Detection in Large Language Models

Jinhua Zhu, Javier Conde, Zhen Gao, Pedro Reviriego, Shanshan Liu and Fabrizio Lombardi

**Abstract**—The utilization of Large Language Models (LLMs) requires dependable operation in the presence of errors in the hardware (caused by for example radiation) as this has become a pressing concern. At the same time, the scale and complexity of LLMs limit the overhead that can be added to detect errors. Therefore, there is a need for low-cost error detection schemes. Concurrent Error Detection (CED) uses the properties of a system to detect errors, so it is an appealing approach. In this paper, we present a new methodology and scheme for error detection in LLMs: Concurrent Linguistic Error Detection (CLED). Its main principle is that the output of LLMs should be valid and generate coherent text; therefore, when the text is not valid or differs significantly from the normal text, it is likely that there is an error. Hence, errors can potentially be detected by checking the linguistic features of the text generated by LLMs. This has the following main advantages: 1) low overhead as the checks are simple and 2) general applicability, so regardless of the LLM implementation details because the text correctness is not related to the LLM algorithms or implementations. The proposed CLED has been evaluated on two LLMs: T5 and OPUS-MT. The results show that with a 1% overhead, CLED can detect more than 87% of the errors, making it suitable to improve LLM dependability at low cost.

## 1 INTRODUCTION

Large language models (LLMs) are becoming a centerpiece of many applications and services; moreover, this trend is expected to continue in the coming years [1]. From the first popular models that had hundreds of millions of parameters such as BERT [2] or T5 [3], models have quickly evolved to billions and in some cases possibly trillions of parameters. Examples of LLMs include commercial models such as GPT [4] or Gemini [5] and open models such as LLama [6] or Mistral [7]. LLMs are also used in safety-critical systems and applications; for example, LLMs are being considered by major car makers for voice command and natural language interactions between the vehicle and passengers [8]. This is just the onset, because LLMs will likely be the default interface of autonomous agents which take actions that can affect persons, hence making the dependability of LLMs a priority going forward. Unfortunately, the large size of LLMs limits the overhead that can be introduced to detect errors, making cost-efficient LLM protection a key issue for their utilization in safety-critical systems.

Protection against errors can be addressed at different levels. For example, for servers or high performance computing units, a significant effort is at the hardware level to detect and correct errors to usually meet stringent Reliability, Availability, and Serviceability (RAS) requirements [9]. Errors can also be managed at software or application level. In that case, a commonly used approach to detect transient errors is to run the same code twice and compare the output results [10]; for LLMs this implies doubling the already large inference time and power dissipation. However, error detection and protection of LLMs at application level have not been widely studied. Many works have considered the impact of errors on neural networks and their protection when implemented on different platforms because this is a key requirement for their use in safety-critical applications [11]. The impact of errors on neural networks has been evaluated extensively both by simulation and by radiation testing [12], [13] showing that they have some intrinsic tolerance to errors, especially when a fixed-point representation is used for the parameters and arithmetic operations. The impact of errors on neural networks implemented on Field Programmable Gate Arrays (FPGAs) has also been considered showing that the implementation of an ensemble of networks can be used to protect against errors [14]. As transformer-based LLMs become widely used in many applications, the next step is to study the impact of soft errors and design error detection and correction schemes to protect the LLMs. The impact of soft errors on BERT has been recently studied in [15] and the results have shown

*This work was supported by the Natural Science Funds of China (62171313), FUN4DATE (PID2022-136684OB-C22) and SMARTY (PCI2024-153434) projects funded by the Spanish Agencia Estatal de Investigacion (AEI) (doi:10.13039/501100011033) and by the European Commission through the Chips Act Joint Undertaking project SMARTY (Grant 101140087). (Corresponding author: Zhen Gao, Shanshan Liu)*

*Jinhua Zhu is now with School of Integrated Circuit Science and Engineering, Tianjin University of Technology, Tianjin 300384, China, and was with School of Electrical and Information Engineering, Tianjin University, Tianjin 300072, China. Email: zhujh@tju.edu.cn.*

*Zhen Gao is with School of Electrical and Information Engineering, Tianjin University, Tianjin 300072, China. Email: zgao@tju.edu.cn.*

*J. Conde and P. Reviriego are with ETSI de Telecomunicacio´n, Universidad Polite´cnica de Madrid, 28040 Madrid, Spain. Email: javier.conde.diaz@upm.es,pedro.reviriego@upm.es.*

*S. Liu is with the University of Electronic Science and Technology of China, Chengdu 611731, Sichuan, China. Email: ssliu@uestc.edu.cn.*

*F. Lombardi is with Northeastern University, Dept. of ECE, Boston, MA 02115, USA. Email: lombardi@ece.neu.edu.*

that even a single error bit flip can in some cases corrupt the output of the entire model when using floating point formats as commonly the case in Graphics Processing Units (GPU) implementations. However, to the best of the authors' knowledge, the impact of soft errors on the parameters of other popular models such as T5 [3] or OPUS-MT [16] and the protection schemes have not been studied.

The detection of errors by using additional logic that operates concurrently with the main system is attractive because it can exploit features of the system for error detection and has been widely used in computing and signal processing systems [17], [18], [19]. Error detection using a concurrent classifier has been proposed for large-scale machine learning systems in [20] and evaluated for BERT in question and answering and emotion classification applications, showing good detection capabilities. However, these schemes rely on having access to internal nodes of the model; this is not always possible for LLMs, for example when using commercial models through an Application Programming Interface (API).

In the case of LLMs when used to produce text, for example in summarization, translation, or conversational tasks, a possible alternative could be to use the properties of the text to detect errors. For example, errors that produce invalid sequences of characters or abnormal patterns could be detected by extracting a few features of the generated text and comparing them with those of normal texts. The key observation is that the output of the LLMs (i.e., the text generated) should have some features that are determined by the rules of the language which may possibly be used to perform concurrent error detection or more precisely, language-based error detection at application level.

In this paper, this is pursued by proposing Concurrent Linguistic Error Detection (CLED), a new methodology to detect errors in LLMs by checking the features of the text generated by the models. The proposed methodology and related scheme are validated on two different models when used in two different applications: T5 when used for summarization and OPUS-MT when used for translation. The results show that CLED enables the concurrent detection of errors at low cost in both cases.

The main contributions of this work are:

1) To study the dependability of T5 and OPUS-MT when they suffer from the presence of a single-bit error on the parameters showing that both models have a similar behavior as other transformer-based models previously evaluated such as BERT [15].
2) To propose and present a new methodology to detect errors in LLMs by exploiting the linguistic features of the text they generate.
3) To evaluate and show the effectiveness of the proposed methodology and scheme in two different models and applications, achieving error detection rates larger than 87% with an overhead lower than 1%.

The rest of this paper is organized as follows. Section 2 covers the preliminaries starting with a summary of related works; it then discusses the error model considered and briefly describes OPUS-MT and T5, as the two language models that are evaluated. The dependability of the OPUS-MT and T5 is studied in Section 3 with error injection experiments that identify the critical bits and the patterns of errors on the generated text. Section 4 presents the proposed CLED describing the scheme by which it can be applied to detect errors in the text generated. CLED is evaluated in Section 5 in terms of error detection effectiveness and the overhead introduced over the unprotected language model. Finally, the paper ends with the conclusion and future work in Section 6.

## 2 PRELIMINARIES

This section covers the preliminaries by first summarizing the most relevant works on techniques to detect errors in LLM and focusing on concurrent error detection (CED). Then the considered error model is discussed and finally briefly describe the language models considered in the evaluation OPUS-MT and T5.

### 2.1 Related Work

The dependability of large-scale machine learning models such as vision transformers [21], diffusion models [22] or LLMs [23] is currently being studied; as new models are introduced, dependability evaluations are also conducted. The results have shown (as for other large machine learning models) an intrinsic tolerance to error(s) in many of the parameter bits; however, there is also the presence of critical bits on which an error can have a large impact on the model [2]. To mitigate the impact of these so-called critical errors protection techniques that modify the implementation have been proposed. For example, as done previously for neural networks [24], by clipping the values of internal signals to avoid bit errors from generating large values that are not seen during normal operation [25].

The use of techniques that modify the implementation of a machine learning model has several issues. The first issue is the requirement to have access to the model, that in many LLMs is not possible. The second issue is that the implementation of protection requires a substantial engineering effort (and time) per model, so for each new model, protection must be implemented again by considering the specificities of the new LLM. Hence in an environment in which new LLMs are constantly being released, the protection to each new model requires a large effort.

An alternative approach, known as CED is to perform error detection concurrently with the system operation using external modules [18]. For example, in the Fast Fourier Transform (FFT), error detection can be performed by checking that the energy of the inputs is the same as the energy of the outputs [26]. This check is independent of the FFT implementation and can be done without making changes to the hardware. Similar schemes that exploit the relations between inputs and outputs, have been proposed for other signal processing blocks [17]. For machine learning systems, the use of a concurrent classifier has been recently proposed [20]. This scheme (referred to as Concurrent Classifier Error Detection (CCED)) relies on monitoring several internal nodes of the machine learning model and feeding their values to a classifier that detects the errors (Fig. 1). The scheme has been evaluated for BERT when it is used for emotion classification or to locate the answer to a question in each text. The nodes monitored in the CCED scheme of [20] were the softmax values, which are used to select the emotion or the position of the answer in the text. The use of internal nodes is applicable to systems that produce outputs with no inherent structure or properties for achieving error detection. For example, in emotion classification, the selected emotion provides little information on whether the system is in error. Instead, the distribution of softmax values among the possible emotions can exhibit different patterns when there is an error, such that the concurrent classifier can detect the error. This is the case for BERT as shown in the evaluation presented in [20].

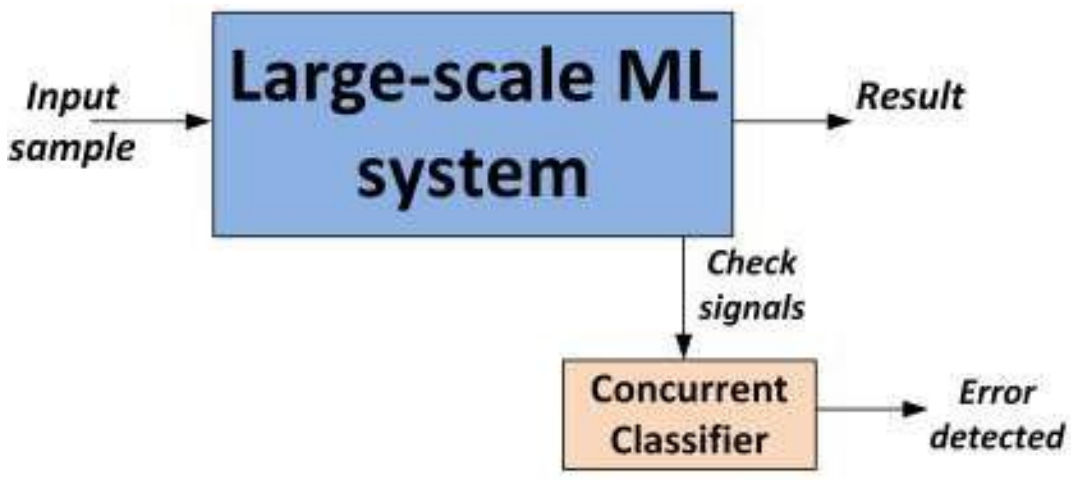

Fig. 1. Block diagram of Concurrent Classifier Error Detection (CCED) [20].

However, when the models are used to generate text (as commonly found in many applications of LLMs), error detection based on the softmax values has some intrinsic limitations. As a first limitation, some of the models are closed and do not expose the softmax values. For example, OpenAI only provides the values of the top five tokens[1] so limiting the use of CCED. A second limitation is that CCED could be significantly more complex because 1) the number of tokens is typically in the order of tens of thousands, so the concurrent classifier will no longer be simple, and 2) it must be run per token so for an interaction with the model, as an example answering a question, hundreds of runs of a more complex concurrent classifier are likely to be needed.

An interesting observation is that when LLMs are used to generate text, their outputs can be used for error detection. This is in line with previous studies on diagnosing and detecting failures from observations of the system's outputs for which general approaches have been proposed [27]. In our case, it means that the CCED scheme can be extended to work directly on the output of the system. Text that is not coherent and consistent is a strong indication of an error; this is somewhat similar to several out of distribution approaches proposed for trustworthiness in neural networks [28]. This makes error detection significantly simpler and more widely applicable to models for which there is no access to internal nodes; moreover, it reduces the complexity and overhead of the error detection process. This is the objective of this paper, to present a methodology for CED in LLMs at application level based solely on the features of the generated text.

### 2.2 Error Model

The objective of the error model is to capture the impact of transient soft errors on the language models. This is motivated by the importance of radiation-induced soft errors in computing components such as memories, GPUs, or FPGAs [29], [30]. Soft errors can affect both sequential logic elements flipping typically one of the bits stored and also combinational logic elements that can propagate the error until a sequential element catches an incorrect value [31]. In the first case, simulation of the errors, for example in the parameters of the model, can be easily performed by flipping one of the bits. However, the simulation of the effect of errors on combinational logic is more complex because it requires access to the detailed design of the component, for example, a GPU on which the model is run. This is typically not feasible and if it were, it would make the results specific to the platform evaluated.

Therefore, based on the practical constraints of simulating errors in combinational logic and to make the evaluation independent of specific components when implementing the LLMs, we consider a single-bit error in one of the LLM parameters. The bit errors are assumed to be equally likely on all parameters and bits, so no assumption is made on the type of parameter affected by the errors. This model captures the effect of soft errors on the model parameters when stored inside a computing unit, for example, in a cache or a register file inside a GPU. The effects of soft errors in combinational elements are not fully captured, but to some extent, we can expect them to be like errors on the model parameters for floating point representations for which errors on the most significant bits of the exponent are critical [15]. Hence, the error model enables us to explore the effects of transient soft errors during the execution of an LLM on a general computing platform.

Since the errors considered in this paper are soft errors, they do not induce any permanent damage to the circuit functionality. We also assume them to occur inside the computational unit (for example a GPU). Therefore, once an error is detected when executing the language model, it can be corrected by running the same input text a second time as that implies that the LLM parameters are reloaded from memory and any previous intermediate value is removed. Therefore, the CLED methodology and scheme proposed in this paper can correct the detected errors by re-computation. Note that, unlike a traditional re-computation that must rerun all input texts, CLED only runs a second time for input texts for which an error has been detected. When the outcome of the error detection process is a false positive, the second run produces the same result, detecting also an error. Therefore, false positives are identified by the detection of an error in the re-computation and have no impact on the application.

### 2.3 LLMs: OPUS-MT and T5

In this subsection the LLMs considered in our evaluation work are briefly reviewed. The first model is OPUS-MT [16] (Fig. 2); it employs the traditional Transformer model with an encoder and a decoder, followed by a prediction layer. The encoder consists of $N$ layers, and each layer is made of two sub-layers, Self-Attention (SA) and Feed-Forward Network (FFN). The decoder also consists of $N$ layers; however, different from the encoder, there is a cross-attention (CA) sub-layer between the SA and the FFN in

1. see https://platform.openai.com/docs/api-reference/chat/create

each layer of the decoder. The prediction layer consists of a linear operation and a log softmax operation. The details of the encoder and decoder are introduced next.

For Natural Language Processing (NLP) applications, the input $X$ of the encoder is a $M \times W$ embedding matrix, where $M$ is the number of input tokens and $W$ is the embedding dimension. In the SA sub-layer, the Multi-Head self-attention (MH-SA) performs the dot-product attention for a query matrix $Q$, a key matrix $K$, and a value matrix $V$. Matrix $Q$, $K$ and $V$ are the linear transformations of $X$, and the weight matrices for the three transformations are $W_Q$, $W_K$, and $W_V$, respectively. As shown in the upper left corner of Fig. 2, the MH-SA mechanism includes $H$ heads, and the outputs of all heads are then concatenated as a $M \times W$ matrix, and merged by an $W \times W$ SA output matrix $W_O$. The Feed Forward (FF) block is composed of two fully connected layers and a non-linear activation GeLU between them, and the weight matrices for the two fully-connected layers are denoted as $W_{F1}$ and $W_{F2}$, respectively. Finally, after residual addition and layer normalization, the output of one layer in the encoder is fed as input to the next encoder layer.

The decoder generates the next token based on the previous predicted tokens. The SA in the decoder is almost the same as in the encoder, but it applies a mask matrix to mask the information after the current token. The processing of the CA is like as SA; the only difference is that $K$ and $V$ are generated based on the encoder output rather than the output of SA. The FFN in the decoder is the same as in the encoder.

The OPUS-MT model includes 6 encoder layers and 6 decoder layers ($N = 6$), and each MH-SA includes 8 heads ($H = 8$). The embedding dimension is $W = 512$. The total number of parameters in OPUS-MT is approximately 78 million, so needing a memory of 156 million bytes when the parameters are stored in half-precision floating-point format.

The second LLM considered in this paper is T5 [3], a larger model that is a variant of the standard Transformer. T5 introduces several improvements to the original transformer, and we introduce the two most significant ones next. First, the layer normalization after the residual addition in SA, CA, or FFN in the standard Transformer is moved to the input of those sub-layers. Second, the FF block in T5 is made of three fully connected layers, and the corresponding weight matrices are denoted as $W_{F1}$, $W_{F2}$ and $W_{FO}$, respectively.

The standard T5 model includes 12 encoder layers and 12 decoder layers ($N = 12$), and each MH-SA includes 12 heads ($H = 12$). The embedding dimension is $W = 768$. Then the total number of parameters in T5 is approximately 247 million, which needs a memory of 494 million bytes for the half-precision floating point format.

## 3 Dependability Evaluation of LLMs

Prior to implementing the CLED scheme, the dependability of T5 and OPUS-MT against single errors on weights is separately evaluated. The objective of this section is to characterize soft errors and their effects on both models for the tasks of summarization and translation, as these soft errors can be detected using CLED directly through the output text. The model performance is measured using traditional metrics that are widely used for each task: ROUGE for summarization and BLEU for machine translation [32]. For each model, we first evaluate the impact of errors on different bit positions of the weights to identify the Critical Bits (CBs) that have a large impact on the model performance, and then analyze the impact of bit flips on the different CBs for the model performance. In this paper, all weights are represented in single-precision floating point (float32). A float32 value consists of a sign bit, 8 exponent bits, and 23 fraction bits, and its value is given by

$$value = (-1)^{sign} \times 2^{exponent-127} \times \left(1 + fraction/2^{23}\right), \quad (1)$$

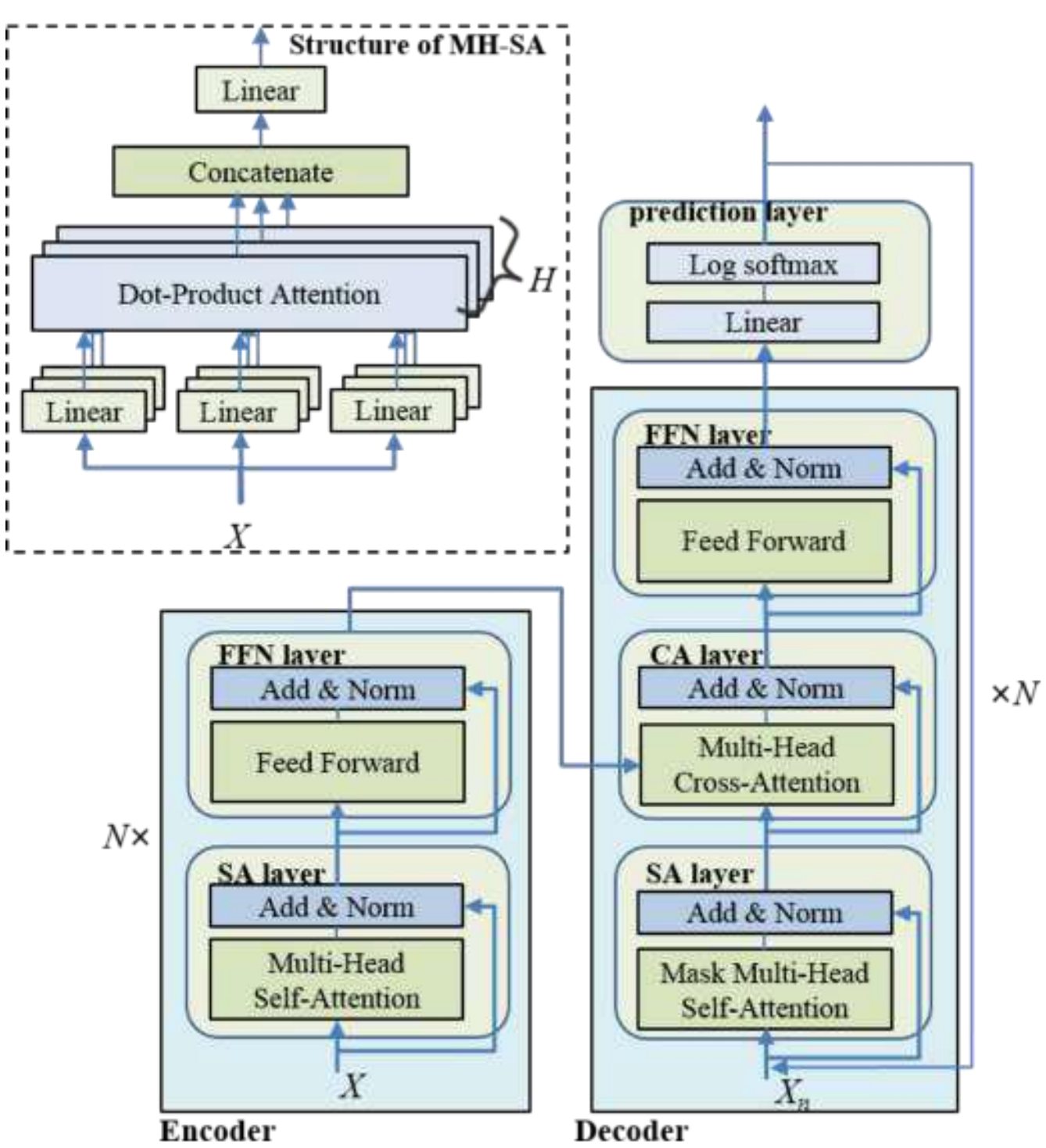

Fig. 2. The OPUS-MT Model.

### 3.1 Dependability Evaluation for T5

1) Dataset and evaluation metric

To evaluate the performance of T5 summarization, the test set of CNN Daily Mail[2] is used and the Rouge1 score is applied as the performance metric, which is calculated as

$$Rouge1 = \frac{\sum_{S \in \{RefSummaries\}} \sum_{gram_1 \in S} Count_{match}(gram_1)}{\sum_{S \in \{RefSummaries\}} \sum_{gram_1 \in S} Count(gram_1)}, \quad (2)$$

where, the denominator is the total number of 1-grams, and the numerator is the number of common 1-grams between the reference summary and generated summary. When there are no errors, the Rouge1 for T5 summarization is 0.4081.

2. Further details on the datasets used for T5 and OPUS-MT are provided in the evaluation section.

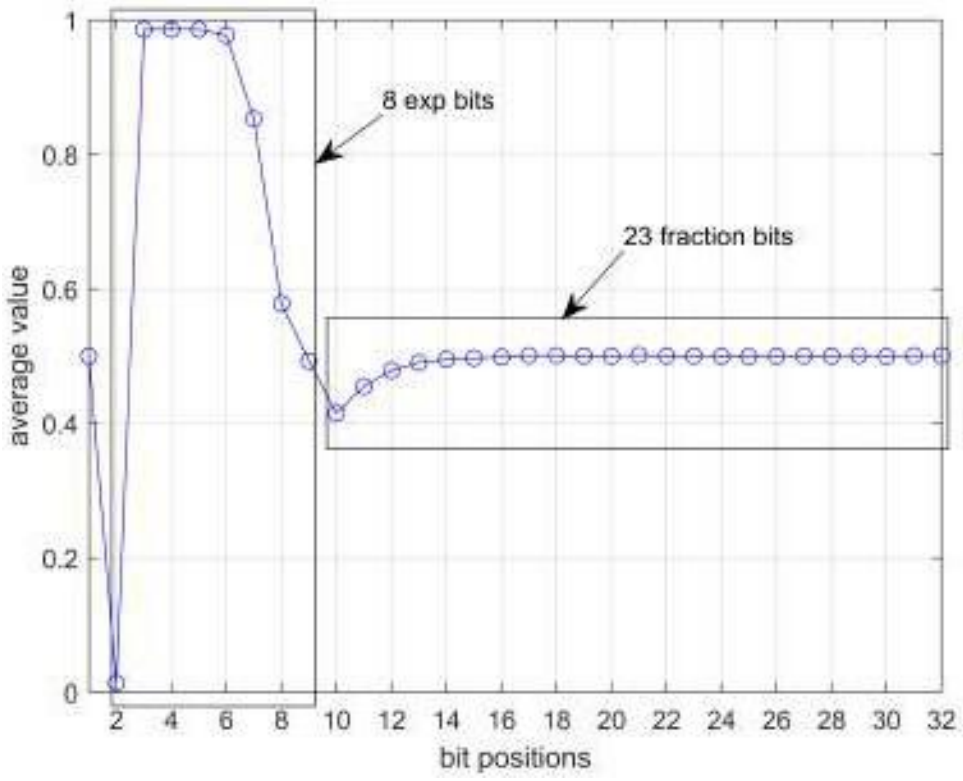

Fig. 3. Average value of each bit for the weights in the T5 model.

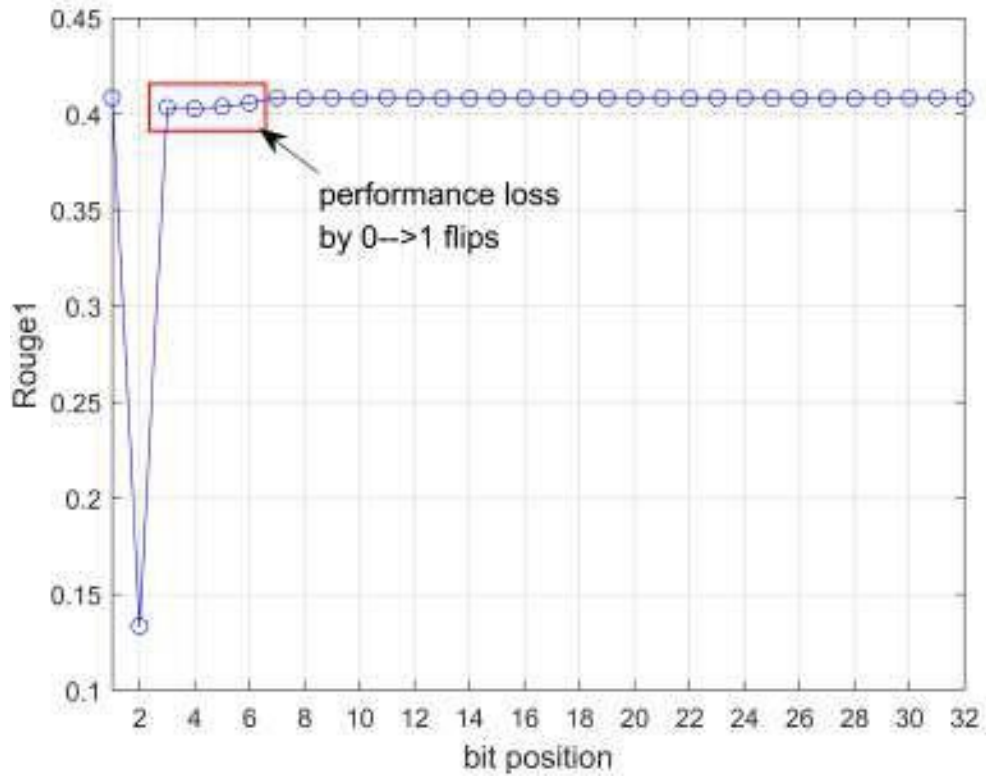

Fig. 4. Impact of errors on different bits of the weights in the T5 model.

2) Distribution of weights in the T5 model

To study the impact of errors on different bits, we first analyze the distribution of all weights in T5 that are publicly available, including $W_Q$, $W_K$, $W_V$ , $W_O$, $W_{F\,i1}$, $W_{F\,i2}$ and $W_{F\,o}$ of the encoder and those weight matrices in the decoder. Based on our analysis, the distribution of the weights for different layers and types is almost the same, and the average value of each bit for single precision weights is plotted in Fig. 3. Based on this figure, we can get three observations: (a) the 1st exponent bit is highly likely to be 0; (b) the 2nd to 6th exponent bits are highly likely to be 1; (c) the average value of other bits is approximately 0.5. This means that the exponents bits have little variability and only the mantissa bits change frequently.

3) General impact of errors on different bit positions

For a floating-point representation, a 0 to 1 flip of the $i$-th exponent bit increases the value of the weight by a factor of $2^{128}/2^{i-1}$, so errors on the upper exponent bits have a larger impact. In this section, we perform error injection experiments to evaluate the effect of flips on different bit positions. In the test for a specific bit position (e.g., $x$), we first randomly select one parameter, flip bit-$x$ of this parameter, and then run the model on the test set. If at least one of the summaries is different from the error-free one, then the error Rouge1 score is recorded. The process is repeated 1000 times, and the average Rouge1 value is used as the significance of bit-$x$. The results are shown in Fig. 4. The flips of the 1st exponent bit introduce a severe performance degradation, and the flips of the 2nd to 5th exponent bits introduce a slight performance loss. All 0 to 1 flips of the first 5 exponent bits cause an obvious performance loss (as shown in Table 1). As the 2nd to 5th exponent bits are 0 for only 1% of the cases, the overall performance loss seems small; however, in general, these results show that the first 5 exponent bits are the CBs for T5.

The experiment results show that in both the encoder and decoder errors on CBs of $W_Q$ and $W_K$ have negligible impact on the model performance due to the softmax operation, and those on the 1st exponent bit of $W_{F1}$ only introduce a few grammar errors in the output summary due to the GeLU operation. Therefore, the performance degradation mainly comes from the 0 to 1 flip of CBs of $W_V$ , $W_O$, $W_{F2}$ and $W_{Fo}$.

4) Analysis of the impact of errors on CBs

TABLE 1
Impact of 0 to 1 errors on the first 5 exponent bits of the weights in the T5 model

| Exp bit | 1 | 2 | 3 | 4 | 5 |
|---|---|---|---|---|---|
| Rouge 1 | 0.0881 | 0.06 | 0.0282 | 0.1042 | 0.3089 |

The effect of errors on the CBs of $W_V$ , $W_O$, $W_{F\,2}$ and $W_{F\,o}$ are similar, and the results for different CBs are listed in Table II, in which the typical wrong outputs are also provided; the error on the 1st exponent bit will result in a fixed output that is irrelevant to the input text. Errors on the 2nd to 4th exponent bits also result in irrelevant summaries, but the outputs change for different inputs; errors on the 5th exponent bit only result in grammar errors in the output text. A discussion of the causes of these error characteristics is presented next.

- Errors on the 1st exponent bit introduce abnormally large values (positive or negative) in the intermediate outputs, which may cause overflow when performing layer normalization. Since the T5 model does not have bias parameters, the overflow causes a 0 output of the layer normalization. Irrespective of which layer the errors occur (encoder or decoder), such 0 outputs force the final results to be fixed and unrelated to the input text.
- Errors on the 2nd to 5th exponent bits do not cause overflow in the layer normalization operation, but the wrong values in the intermediate results cause changes in the final logits scores for the word selection in the prediction layer. For errors on the 2nd, 3rd, and 4th exponent bits of some large weights, the logits scores change significantly, so that the output is irrelevant to the input text and looks random. However, for the 4th exponent bits of some small weights and the 5th exponent bits of the weights, errors only introduce small changes to the logits scores, which would then cause some grammar errors in the final output. For example, the errors on the 5th exponent bit may cause repeating words in the text as shown in Table 2.

### 3.2 Dependability Evaluation for OPUS MT

1) Dataset and evaluation metric

TABLE 2
Effect of errors on different CBs for T5 based summary

| Bit position | Error characteristics | Typical output examples |
|---|---|---|
| 1 | Fixed sentence or strings | Decoder: the the thea the the: the thet thesss thes the thesassa thes:ss: (omit subsequent characters) Encoder: iReporter.com: What's New in New York City? . .: What Happens Next? - What Happened To You? "It's a Wonderful Life" |
| 2 | Random and similar strings | age .snaes-drubs of swaas - saber sdberbersedst syscercerssedcersssce of eraskar a yror .age . |
| 3 | Random and similar strings | distr bebelus adanc spalat intrebgardinen Clickfunnel mes insotit |
| 4 | Random and similar strings or grammar errors | agments cans, a symgael, is a noggottadfad, he says . Previously he and ad samess, the duddad isn't ay |
| 5 | Grammar errors | The Palestinian Authority signed the International Criminal Criminal on Wednesday (omit subsequent characters) |

We use the IWSLT2017 dataset [33] in this section, and the BLEU score is used to evaluate the translation performance of OPUS-MT, which is defined as

$$\begin{cases} BLEU = BP \cdot \exp\left(\sum_{n=1}^{N} w_n \cdot \log p_n\right) \\ BP = \begin{cases} 1, & c > r \\ \exp(1 - r/c), & c \le r \end{cases}, \\ p_n = \dfrac{\min(count^n_{match}, count^n_{ref})}{count^n_{output}}, \end{cases} \tag{3}$$

where $p_n$ is the accuracy of $n$-grams, $count^n_{output}$ and $count^n_{ref}$ represent the number of $n$-grams in the output translation and the reference translation, respectively, $count^n_{match}$ is the number of common $n$-grams in both the output translation and the reference translation, $BP$ is the penalty factor, $c$ is the sentence length of the output translation, and $r$ is the sentence length of the reference translation. When there are no errors, the BLEU score is 0.2186.

2) Identification of CBs for OPUS-MT model

The weight distribution of the OPUS-MT model is first analyzed, and the average values of different bits of the weights are shown in Fig. 5. As we can see, the results are similar to T5, but they also reveal two differences: (a) the 1st exponent bit is always 0; (b) the 2nd to 4th exponent bits are always 1.

Then we perform error injection experiments to evaluate the impact of errors on different bit positions, and the average BLEU scores over 1000 runs are shown in Fig. 5. Only errors on the 1st exponent bit introduce a severe performance loss, while those on other bits do not affect the model performance. These results are expected because errors on the (upper) exponent bits lower than the 1st bit only decrease the weights, and thus, they do not affect the model performance. Therefore, the 1st exponent bit is the only CB for the OPUS-MT model.

3) Analysis of the impact of errors on the CB

Similar to the results for T5, the effects on errors on $W_Q$, $W_K$ and $W_{f1}$ are reduced by softmax and GeLU, the performance loss mainly comes from the errors on $W_V$, $W_O$ and $W_{f2}$. However, different from the cases for T5, the layer normalization in OPUS-MT also includes bias, so the layer output becomes a fixed matrix (with each row equal to the bias vector) instead of a 0 matrix. Therefore, if the errors occur in the encoder part, the translation results are determined by the bias vector of the layer norm, and they are not related to the input text. Since the layer norm bias is different for different layers, the wrong translation results also change for different layers, but it is fixed for different inputs. The errors in a layer of the decoder part also produce a fixed layer output matrix with each row becoming the bias vector, and finally generating a wrong translation result that is irrelevant to the input text. However, a difference is encountered, namely the fixed layer output is only used for the computation of $Q$ of the next layer, and the computations for $K$ and $V$ of the following layers are based on the correct results from the encoder. Therefore, the final results are different for different inputs. Table 3 shows some examples of the wrong translation results for different layers in the encoder and decoder parts. The results are meaningless and include a repeat of fixed words or symbols. This provides the reasons for the BLEU score for errors on the 1st exponent bit being 0 in Fig. 6.

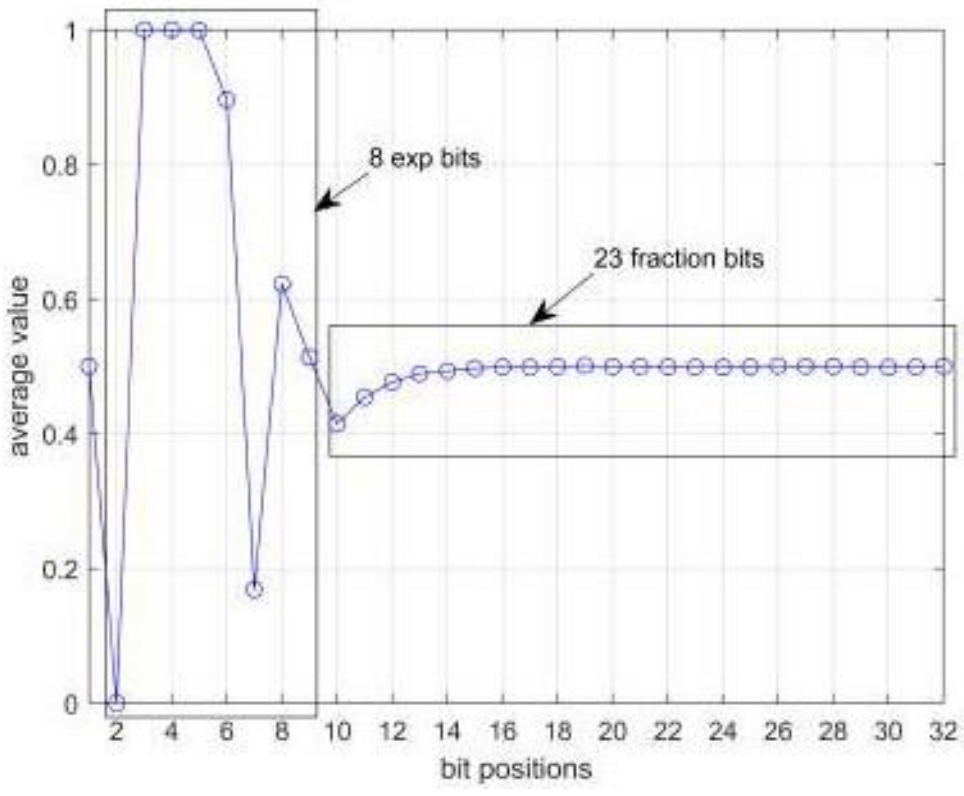

Fig. 5. Average value of each bit for OPUS-MT model.

TABLE 3
Effect of errors on the CB for OPUS-MT based translation

| | Layer | Translation result (the first line) |
|---|---|---|
| Encoder | 1 | All-for- and-for-for- and-for-for- and-for-for- and-for- |
| | 3 | I'd be my I——————————— |
| | 5 | It's—'——————————— |
| Decoder | 1 | (empty strings) |
| | 3 | ........................................................................ |
| | 5 | at at at at at at at at at at at at at at at at at at at at at at |

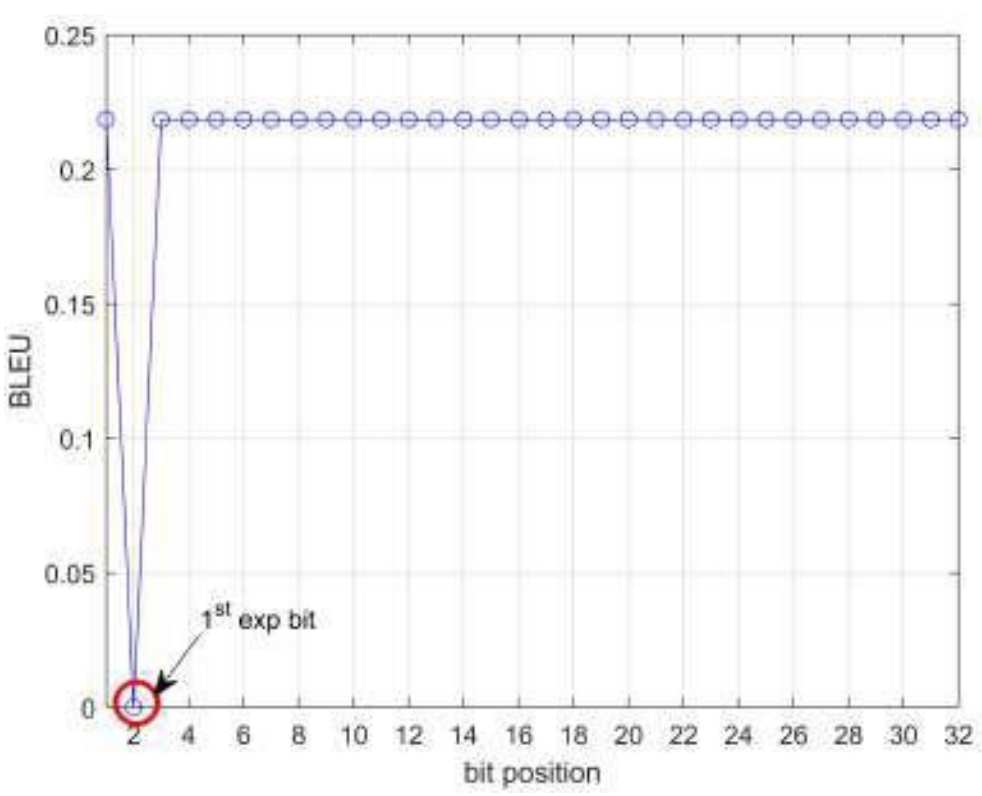

Fig. 6. Impact of errors on each bit for OPUS-MT model.

## 4 Concurrent Linguistic Error Detection (CLED)

In the previous section, the impact of the change of a bit in OPUS-MT and T5 has been analyzed. Error detection is based on the comparison with the error-free output. In this section, the proposed CLED scheme is presented. First, the motivation and high-level approach are presented to then discuss the main blocks of the scheme: the linguistic features and the concurrent classifier.

### 4.1 Motivation

As discussed in prior sections, the main motivation for the development of the proposed CLED methodology and scheme is to provide error detection mechanisms for LLMs that:

1) Operate on models for which there is no access to the internal nodes.
2) Are widely applicable to language models regardless of their implementation details.

Error detection mechanisms with those features can be easily integrated into an LLM workflow such that error detection is just an additional step to be performed during the LLM design, or at a later stage by a different company or organization. This provides both flexibility and scalability to protect LLMs; however, to the best of the authors' knowledge, there is no CED scheme in the technical literature that meets these requirements, the CCED proposed in [20] meets partially 2) but it does not fulfill 1).

### 4.2 Overall Approach

As discussed in the introduction, a trivial but at the same time key observation is that when language models are used to generate text, we expect the text to be valid. This implies that words must be correct, and the sentences and constructions must follow the rules of the language. Intuitively, errors that cause the model to malfunction are likely to output meaningless text or at least incorrect text as discussed in the previous section. As an example, this is an excerpt of the output of T5 with an error injected in one of its parameters: "Satoisesaias,mass of a customiseaed toeasaiesaado hudness in a shader in hiuued tuuuiut tiuisuaiui"; it can be seen that the text is clearly incorrect. This example and the ones presented in the previous section suggest that it may be possible to detect errors based only on the features of the output text to enable the protection of closed models and be applicable to models regardless of their implementation details. This leads to two research questions that will be answered in the rest of this paper:

- RQ1: Is it possible to efficiently detect errors in the text generated by LLMs using linguistic features?
- RQ2: If it is possible to perform error detection based on linguistic features, would it generally apply to different LLMs and use cases?

To address these questions, we start by discussing the proposed CLED scheme illustrated in Fig. 7. It operates on the output text of the LLM and therefore, it does not require any modifications to the model, or access to any internal nodes. This makes CLED attractive to protect models that are provided as a black box to which no modification can be made. The error detection is performed in two steps.

Initially, relevant linguistic features are extracted from the output text and fed to another classifier that performs error detection. Since error detection is based solely on the properties of the text, the approach should be applicable to newer LLMs. This is very interesting as the pace at which models are introduced is hectic and having an error detection scheme that is general and can be applied to any LLM facilitates the fast protection of LLMs as they are introduced. The main elements of the proposed CLED scheme are the linguistic features extractor and the concurrent classifier, both are briefly discussed in the next subsections after presenting the datasets used in the presented evaluation.

The overheads introduced by the proposed scheme for error detection are thus from the extraction of the linguistic features and the concurrent classifier. Given the large size of LLMs, the relative burden of the linguistic features extractor and the concurrent classifiers should be negligible as shown next.

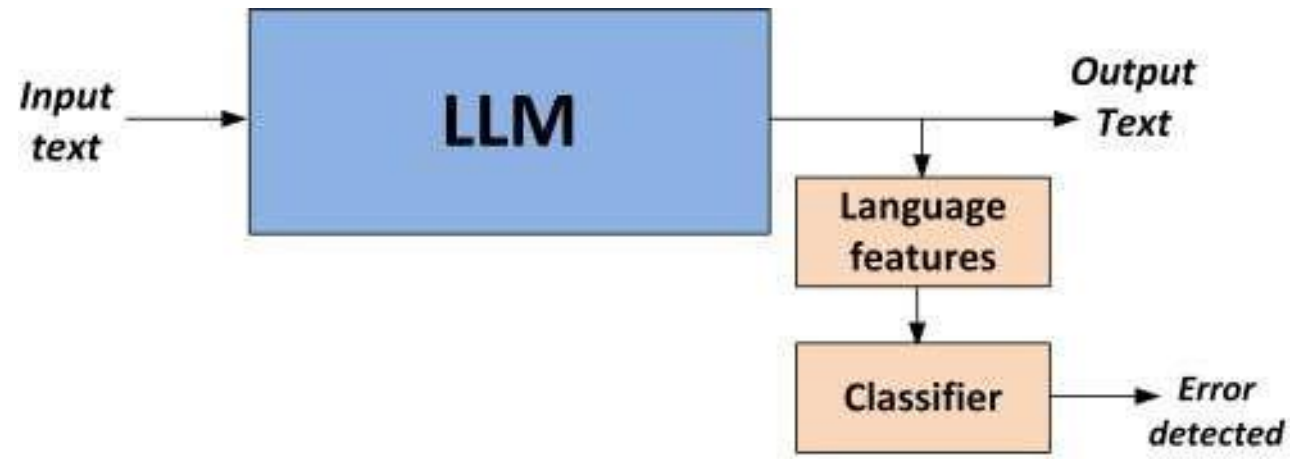

Fig. 7. Proposed Concurrent Linguistic Error Detection (CLED) scheme.

### 4.3 Dataset Description

We tested our proposal using two different datasets. The CNN Daily Mail dataset[3] was used to generate error-free and error-injected summaries with T5, and the IWSLT17 translation dataset [33] was used by applying the same methodology but in the translation task using OPUS-MT.

For the T5 model, a subset of 11,490 news from the CNN Daily Mail dataset has been selected and run to generate

3. CNN Daily Mail dataset: https://huggingface.co/datasets/cnn _ dailymail/viewer/3.0.0/test

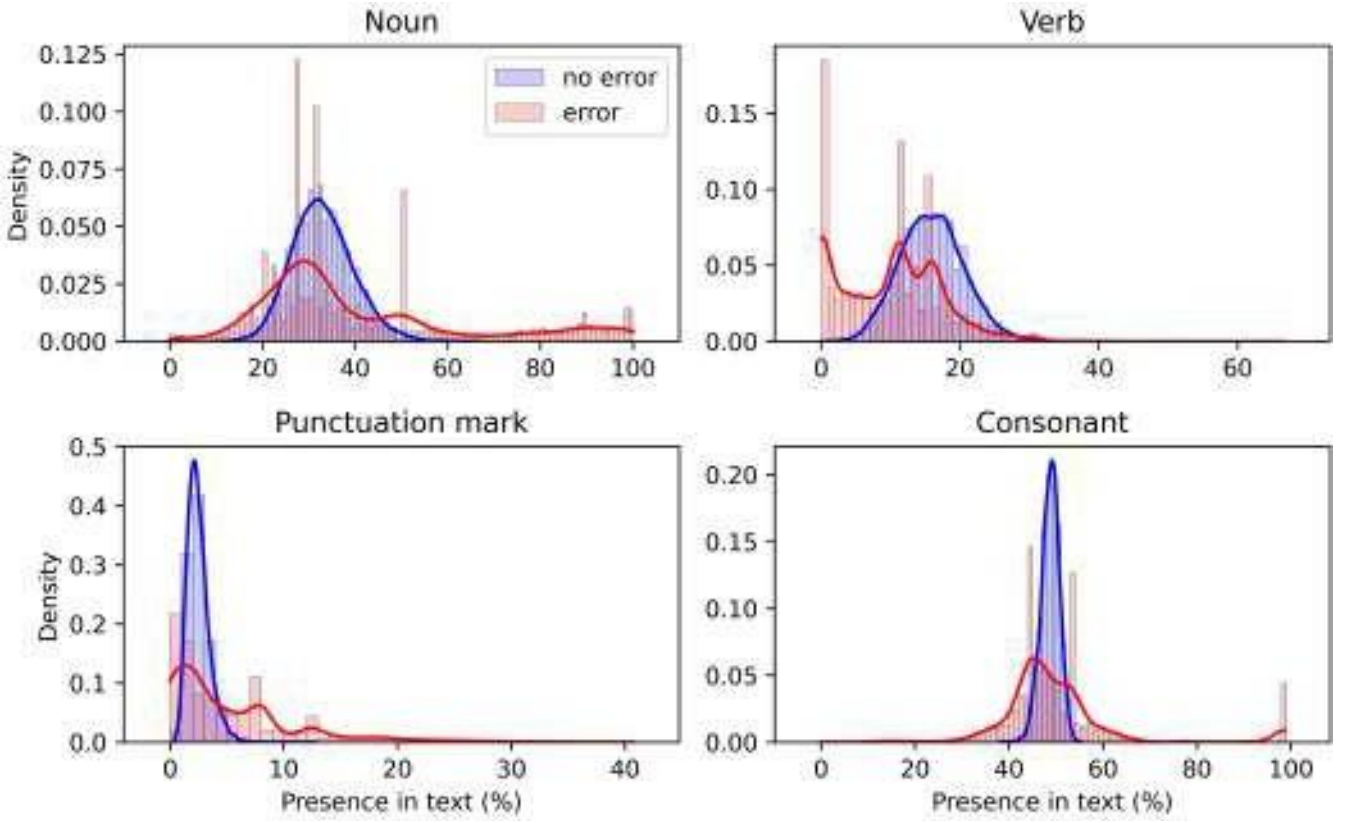

Fig. 8. Comparison of distributions between texts generated by error free model and in the presence of soft errors for T5.

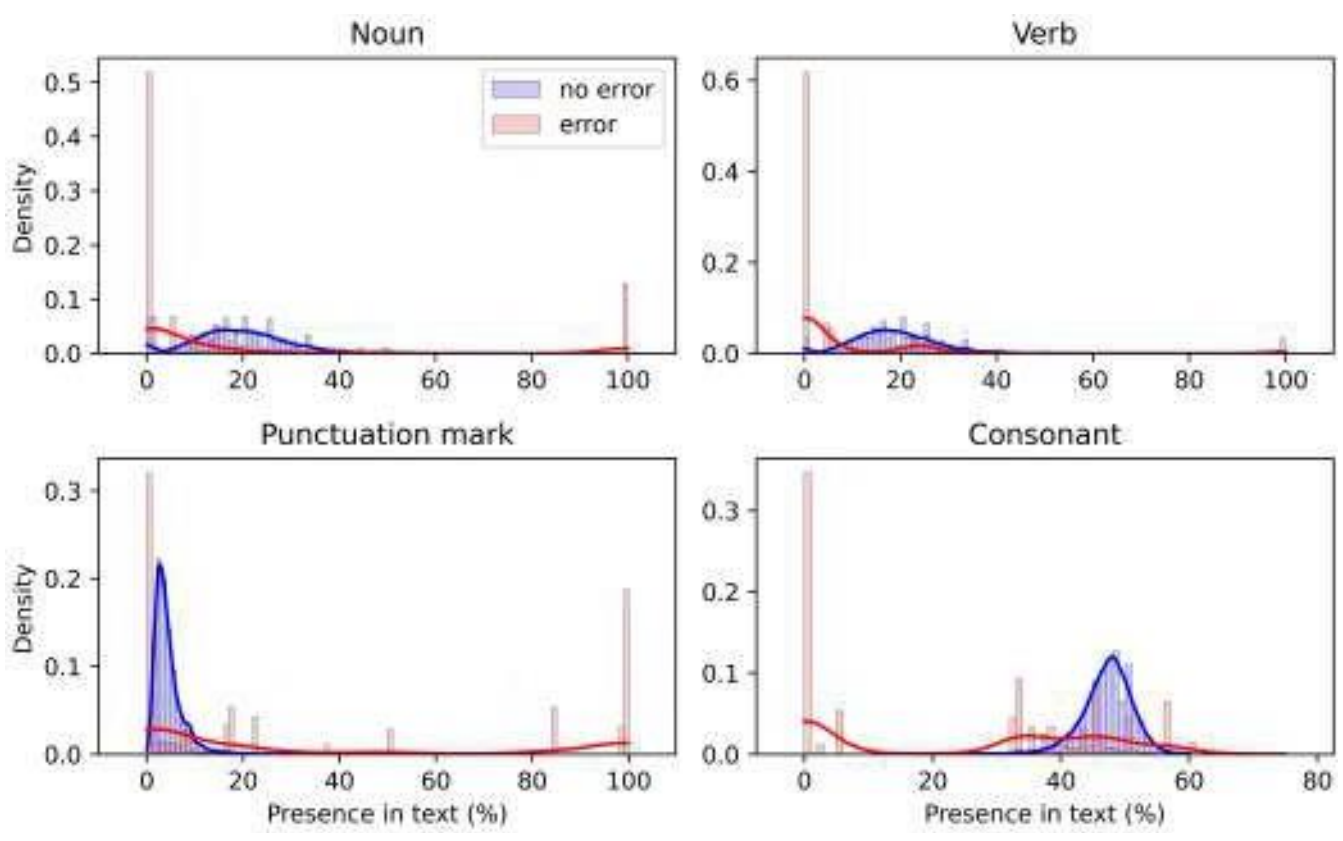

Fig. 9. Comparison of distributions between texts generated by error-free model and in the presence of soft errors for OPUS-MT.

error-free summaries. Then an error has been injected in a randomly selected bit of a randomly selected parameter of the T5 model and the summaries for the same 11,490 news have been generated. If at least one of them was different from the error-free summary, then the error was considered relevant, otherwise if the error had no effect on any of the summaries, the error is assumed to be irrelevant[4]. The process was repeated until 100 relevant errors were identified. Then a dataset has been built with the 11,490 error-free summaries and 11,490 erroneous summaries from 100 relevant errors. The dataset is available online[5]. These error-free and erroneous texts are used as the dataset on which the concurrent classifier is trained and evaluated. The 22,980 samples are split, with 80% allocated for training, and 20% for testing.

A descriptive analysis of the training dataset shows that the differences in the Rouge1 metric observed previously, as well as the patterns for the generation of the errors detected, are also reflected in the lexical characteristics of the texts. Fig. 8 shows the probability distribution of texts summarized by T5 with injected errors, and the texts summarized by the error-free T5 configuration for different categories of words or lexical structures. The error-free texts follow normal distributions in all four cases (nouns, verbs, punctuation marks, and consonants). However, texts with errors do not fit a normal distribution; moreover, a significant number of texts do not contain verbs or punctuation marks, while there are texts composed only of consonants or nouns. These characteristics are utilized in the CLED scheme to detect errors in the model.

For the OPUS-MT model, we followed a similar approach but with a different task. In this scenario the IWSLT17 translation dataset has been used to select 8,549 samples of Chinese-to-English translation and OPUS-MT has been first run with no errors. Then the same process as for T5 was performed to identify 100 relevant errors and build a dataset with 8,459 error-free and 8,459 erroneous samples. The 17,098 samples are split, with 80% allocated for training, and 20% for testing. Again, the dataset is published online[6].

As with T5, errors are also observed in the distributions of the generated texts. In this case, the effects are similar; while the texts translated by the error-free model follow normal distributions, the texts generated by the model with errors no longer exhibit this characteristic (Fig. 9). Many of these texts are characterized by the absence of nouns, verbs, punctuation marks, or consonants. Moreover, in some cases the opposite occurs, and they are made of only nouns, verbs, or punctuation marks. Compared to T5, the distributions differ more in relation to the texts generated by the error-free model. This initial analysis supports the feasibility of detecting errors using linguistic features, so addressing RQ1.

### 4.4 Linguistic Features

In this subsection, we briefly discuss linguistic features that can be used for error detection based on previous results and propose a group of such features for the initial evaluation of CLED. Such a list and selection of features does not intend to be exhaustive nor optimal; the goal is just to illustrate the potential of CLED, and better features can probably be found. For example, the understanding of the way the models operate [34] could provide information on the best features to use for error detection.

These additional improvements are left for future work.

Languages have spelling rules that restrict the possible characteristics of a valid text. In English, for example, the same consonant cannot be tripled, digraphs such as ‘uu’ and ‘aa’ are extremely rare, and ‘ii’ never occurs [35] (pp. 132 and 449). Similarly, in Spanish, only six consonants can appear in a word’s final position [36], and in German, sequences of fricatives (such as /vz/) are not allowed [37]. These rules can potentially be used to detect invalid texts that can be a sign of an error, because the models have been trained to produce valid texts. This seems to be confirmed by looking at the examples of texts generated by the models with an error shown in Tables 2 and 3, invalid words are provided as well as character sequences in many cases.

4. This is commonly the case when the error changes one of the least significant bits of a parameter.

5. Single-bit error dataset: https://doi.org/10.5281/zenodo.10647624

6. Single-bit error dataset: https://doi.org/10.5281/zenodo.10647624

The presence of invalid words is not the only linguistic information that can be used to detect errors. There are also patterns that valid texts follow, for example, the distribution of the part of speech words [38] is typically within some ranges, and not all words are for example, nouns or verbs. These types of patterns can also be used to detect anomalous text caused by errors. Some works have proposed grammaticality when analyzing sentences to detect errors; for example, they extract the grammatical category of words with Part-of-Speech (POS) techniques and train models using the sequences of grammatical categories as features [39]. Other works demonstrate that the frequency of grammatical categories allows training simple classifiers with high accuracy, such as the one developed by Mendhakar and Darshan to differentiate between fiction and non-fiction texts [40]. However, these methods are not capable of detecting semantic errors and can classify sentences like “I play basketball” (correct) as well as “I play melon” (incorrect) as valid. This highlights the inadequacy of probabilistic grammatical models to detect semantic incongruencies, a problem already described by linguists [41].

Therefore, based on the previous discussion, two types of features are used as inputs to the concurrent classifier of the proposed CLED scheme: linguistic rules and linguistic patterns. The first type of feature checks the restrictions that are defined by the language. The second type of feature captures patterns that are common in the language, for example, a text in which the relative frequency of consonants is ten times larger than the number of vowels is also likely to be invalid due to an error. These features used can be adapted to better capture the impact of errors on the output text; some examples of features are:

1) Restriction: number of capital letters in the middle of words.
2) Restriction: number of sequences of the same letter repeating three or more times.
3) Pattern: average word length.
4) Pattern: frequency of vowels or consonants.
5) Pattern: distribution of grammatical categories (i.e., verbs, nouns, determiners, conjunctions.).

Ideally, a common set of features can be used for different LLMs and applications but depending on the performance of the classifier, the features to use may have to be adjusted to better capture the error patterns caused by the soft error in a specific LLM and application. In the proposed design, the same set of features is used for both models (OPUS-MT and T5) and applications (translation and summarization) to illustrate the general applicability of the linguistic features. This facilitates the protection of LLMs because it makes feature extraction independent of the model application. This selection is also intended to address RQ2 on the general applicability of the proposed CLED methodology and scheme.

Table 4 shows the linguistic features used for the concurrent classifier. The first three are rules as they correspond to letter sequences that are invalid while the rest of the features correspond to patterns that are expected to be within a range of values in valid texts. As discussed before, these features will be used both for OPUS-MT and T5 to illustrate that a common set of features can be used for different models and applications.

TABLE 4
Linguistic Features used for the concurrent classifier

| | Feature | Description |
|---|---|---|
| Rule | uppercase_middle | Frequency of words with uppercase in the middle of the word |
| | four_or_more_consonants | Frequency of words with sequences of four or more consonants |
| | three_or_more_eq_chars | Frequency of words with sequences of three or more characters equal |
| Pattern | punctuation_mark | Frequency of punctuation marks in the text |
| | digit | Frequency of digits in the text |
| | blank | Frequency of blanks in the text |
| | vowel | Frequency of vowels in the text |
| | words density | Inverse of average word length |
| | ADP | Frequency of adpositions in the text |
| | NUM | Frequency of numerals in the text |
| | VERB | Frequency of verbs in the text |
| | DET | Frequency of determiners in the text |
| | PRON | Frequency of pronouns in the text |
| | NOUN | Frequency of nouns in the text |
| | PRT | Frequency of particles in the text |
| | ADV | Frequency of adverbs in the text |

### 4.5 Concurrent Classifier

In this section the implementation of the concurrent classifier is discussed. As with the language features, our goal is not to find the optimum classifier, but rather to show the effectiveness of CLED even when using a simple classifier.

As the number of linguistic features is small and they are also simple, a basic binary classifier should be sufficient to detect the errors. In the proposed design, a Random Forest classifier[7] is used as in [20]. Different combinations of models and features have been tested. The model’s hyperparameters have been configured using cross-validation and recall as the metric to evaluate the model to reduce false negatives. This process has been performed independently for OPUS-MT and T5; the same set of linguistic features and the same type of classifier (a Random Forest) is used for both models but trained with the relevant dataset in each case.

### 4.6 Error Correction based on CLED

The proposed CLED can detect errors, however in many applications it is not enough; once an error is detected it should be corrected. In our case as we consider soft errors that do not affect the computational component’s functionality, the model is run a second time for the same input text. This implies that the LLM parameters are reloaded from external memory and previous results, or intermediate values are removed. This process is illustrated in Figure 10. Unless there is a second soft error (which is not likely as soft errors are not frequent events), the occurrence of two errors on consecutive runs is extremely unlikely, i.e., the second execution is usually error-free. Therefore, if CLED detects an error on the second run, it is assumed that it is a false positive and hence, the generated text is assumed to be correct; this is as in [20].

7. https://scikit-learn.org/1.3/modules/generated/sklearn.ensemble.RandomForestClassifier.html

The CLED-based error correction introduces an overhead in terms of LLM re-computations that is given by the false positive rate. For example, if the concurrent classifier falsely detects an error in 1% of the error-free cases, a 1% overhead is incurred for re-computation. This enables a trade-off between error detection and re-computations which is discussed in the next section.

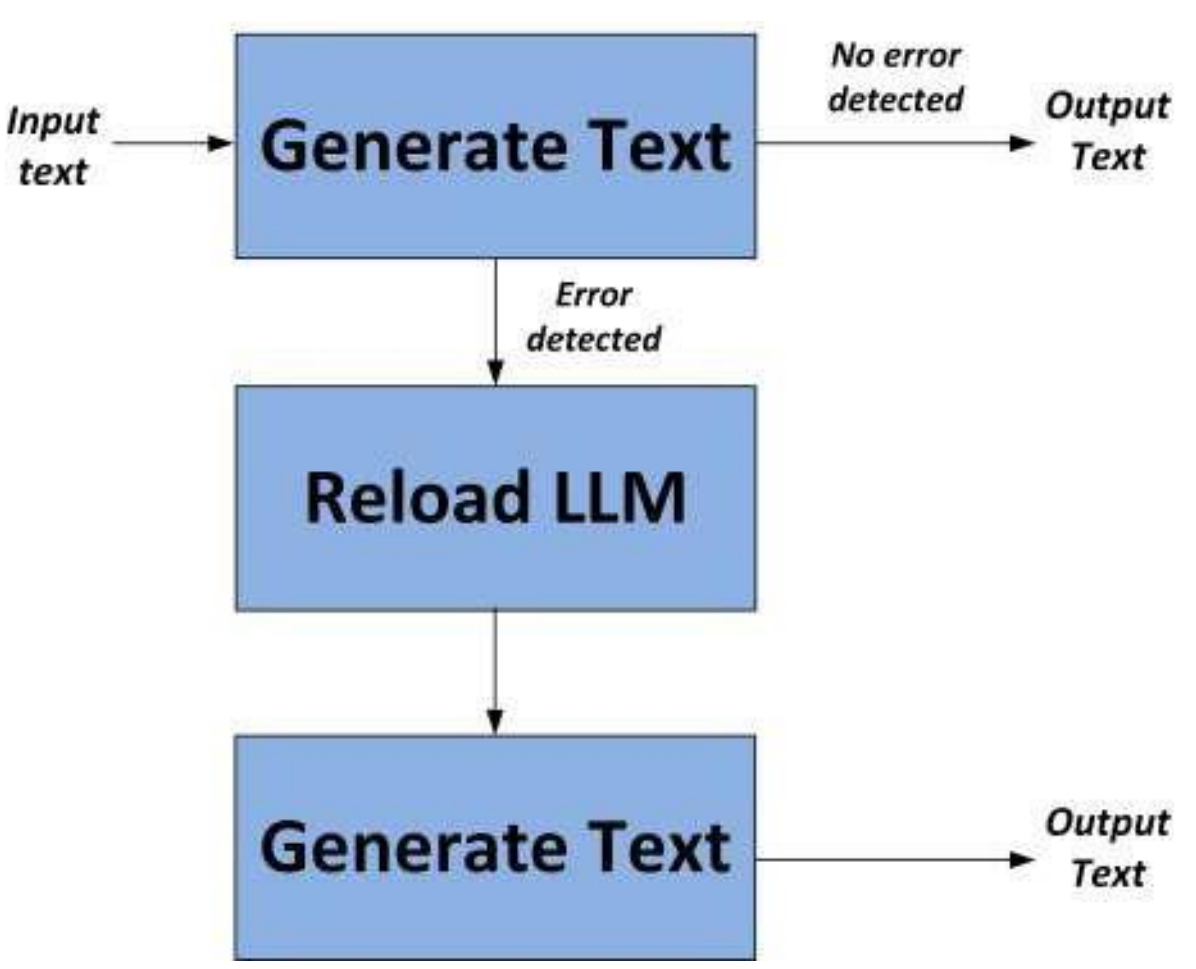

Fig. 10. Error Correction using the Proposed Concurrent Linguistic Error Detection (CLED) scheme.

## 5 EVALUATION

To evaluate the proposed CLED scheme, T5 is applied to text summarization and OPUS-MT is used for Chinese to English translation as case studies. Both models are executed on a GPU (NVIDIA GeForce RTX 4080) and the concurrent classifier on a standard CPU (Intel Xeon E5-2630 v2 @2.6GHZ and 32GB RAM). Errors are injected on a bit of a random parameter prior to executing the models. As discussed previously, CLED targets the protection of LLMs that are seen by the designer as a black box, i.e., with no access to the internal nodes. In this scenario, protection techniques that are applicable and comparable are duplication or re-computation that imply at least doubling the overhead of an implementation.

Next, the results for each of the models and applications are presented and briefly discussed, then the use of CLED based on those results in different scenarios is also presented. The same CLED scheme is also tested on the quantized versions of the models. Many LLMs perform post-training quantization that compresses the model into versions with lower precision in their parameters, so validating CLED in this quantized setting would simplify its adoption and extensibility.

### 5.1 T5 Summarization

The T5 model is used to summarize news taken from the CNN Daily Mail dataset with the free and injected-error configurations of the LLM, generating the dataset described in Section 4.3. These error-free and erroneous texts are used as the dataset on which the concurrent classifier is trained and evaluated. The dataset contains 11,490 error-free summaries and 11,490 erroneous summaries obtained from 100 relevant error configurations. The 22,980 samples are split, with 80% allocated for training, and 20% for testing.

TABLE 5
Model Hyperparameters of the Concurrent Random Forest Classifier

| Hyperparameter | Value | Description |
|---|---|---|
| Number of trees | 50 | Improves the generalization of the model and avoids inter-tree overfitting by limiting the number of trees |
| Maximum depth | 10 | Avoids in-tree overfitting by limiting the maximum number of levels per tree |
| Maximum number of features per split | sqrt | Prevents the selection of a subset of features that may be too large in each split by limiting it to the square root of the number of the features |
| Impurity criterion | Gini | Impurity of a tree based on the probability to incorrectly classify a randomly chosen element if it were randomly labeled |
| Bootstrapping of samples | Yes | Introduces variability in the training datasets of each tree, which contributes to reducing the correlation between the trees |
| Minimum samples per leaf | 2 | Avoids overfitting by requiring at least two samples per leaf in each tree |
| Minimum samples for a node split | 10 | Avoids overfitting by requiring ten samples to make a new split in the tree |
| Maximum leaf nodes | No limit | Adds flexibility to each tree as other parameters limit the overfitting |
| Minimum decrease impurity for splitting | No limit | Allows a new splitting when impurity decreases at any value. Other parameters are fixed to avoid overfitting |
| Minimum weight fraction per leaf | No limit | Allows leaf nodes with any weight fraction value. Other parameters are fixed to avoid overfitting |

Different classifiers with multiple hyperparameters configurations have been tested. The best model has been achieved with a Random Forest comprised of 50 trees maximum depth of 10 levels for each tree, with the square root of the number of features as the maximum number of features considered for each split, gini as the impurity criterion for splitting, with bootstrapping of samples, a minimum of 2 samples per leaf, and requiring at least 10 samples for a node split. Table 5 shows the resulting hyperparameters used to train a model that fits the error detection problem, with the mentioned features, and that avoids overfitting. The results on the test dataset reveal an accuracy of 93% and a recall of 93% with a false negative rate of 11% and a false positive rate of 2%.

Fig. 11 shows the contribution of each feature to the Random Forest classifier; in this case, the punctuation marks are the most decisive feature (importance of 0.18), followed by the frequency of digits (0.12) and adpositions (0.12). This is aligned with the analysis of error effects, that it noted random and similar strings frequently appeared (Section 3.1).

The relative complexity of the main model and the concurrent classifier is evaluated in terms of their average run time per sample in the dataset. The results are summarized in Table 6 where the times for the concurrent classifier include the linguistic feature extraction and the Random Forest classifier. The overhead for concurrent error detection is small even when considering that the main model was run on a GPU (NVIDIA GeForce RTX 4080) and the concurrent classifier on a standard CPU (Intel Xeon E5-2630 v2 @2.6GHZ and 32GB RAM). The runtime for the concurrent classifier is at least one order of magnitude below the main model. It is interesting to note that the complexity of the main model increases continuously, for example, most state-of-the-art LLMs have at least several billion parameters, so significantly more than OPUS-MT and T5. Instead, the complexity of the proposed CLED is independent of the LLM size as it operates directly on the output text. Therefore, the relative overhead of the concurrent classifier will be even lower for more advanced LLMs.

TABLE 6
Average run time for case studies (milliseconds)

| Case study | Main model | Concurrent Classifier |
|---|---|---|
| T5 summarization | 453ms/sample | 11ms/sample |
| OPUS-MT translation | 113ms/sample | 11ms/sample |

Finally, the same classifier was applied to a quantized floating-point 16 (float16) version, which reduces the LLM size by approximately half. The results show a slight decrease in performance, with an accuracy of 90%, a recall of 90%, 2% false positives, and 18% false negatives. However, a classifier trained specifically for the float16 model, using the same features and hyperparameters as the base model, achieves results comparable to the original, with an accuracy of 92%, a recall of 92%, 5% false positives, and 10% false negatives.

## 5.2 OPUS-MT Translation

As a second case study, a different model (OPUS-MT) and application (translation) has been considered. The dataset described in Section 4.3 generated from the IWSLT17 translation dataset [33] has been used in this scenario. The dataset contains 8,459 error-free and 8,459 erroneous samples. The 17,098 samples are split, with 80% allocated for training, and 20% for testing.

Similarly to T5, a Random Forest binary classifier has been trained using the same features (Table 4) and hyperparameter configuration. As a result, a model with an accuracy of 95% and a recall of 95% was obtained, with a false negative rate of 9% and a false positive rate of 1%. Fig. 12 shows the error detection rate for different overhead rates (false positives).

In this case, the frequency of verb occurrence (importance of 0.20), punctuation marks (0.19), and nouns (0.16) are the most important features of the classifier (Fig. 11). This once again aligns with the description of errors detected in OPUS-MT (Section 3.2), where it was observed that with injected-errors the LLM tends to repeat punctuation marks and words. It is interesting to note that the presence of punctuation marks is a common feature for both OPUS-MT and T5, because many of the texts generated with errors are characterized by having either none, or numerous punctuation marks.

The results demonstrate that the use of Random Forest and the selected features enable the training of a concurrent classifier for error detection with high accuracy and low false negative rate for different LLMs and applications. This confirms our intuition that the linguistic features can to some extent be general and used for different LLMs which is the second research question (RQ2) considered in this paper. As previously discussed, this makes CLED more attractive as the feature extraction and classifier can be the same and when changing LLM or application, the designer only needs to train the concurrent classifier with the relevant dataset to achieve effective protection.

Finally, it is worth noting that this task is multilingual which shows the ability of CLED to detect errors independently of the language. This can be attributed to the selection of the features in Table 4 which are mostly language-independent for languages that use a letter script.

As for OPUS-MT, the classifier evaluation was repeated using a model quantized to float16. The results were like those obtained with T5. When using the classifier trained with float32, performance decreased slightly, with an accuracy of 94%, a recall of 94%, 1% false positives, and 12% false negatives. However, as before, training a classifier specifically for float16, using the same features and hyperparameters as in float32, produced results comparable to the original classifier with 96% of accuracy, 96% of recall, 2% false positives, and 5% false negatives.

## 5.3 Detection versus Re-computation Trade-off

To better understand the potential use of the proposed CLED in different applications, the relation between the error detection rate and the false positive rate for both T5 and OPUS-MT is investigated and shown in Fig. 12. A change of the decision threshold of the model allows for control of the overhead of re-computation, i.e., unnecessary re-computations run when an error has been detected but the prediction was wrong (measured by the false positive rate). Moreover, it demonstrates the influence of this re-computation rate on error non-detection, quantified as false negatives.

The results show that most of the errors, close to 90%, can be detected even with a very low, for example, 1%, re-computation overhead. The overhead of detecting additional errors grows significantly and to detect an additional 4% of the errors, the re-computations increase by more than 5 times. Therefore, the designer can trade off detection effectiveness for re-computations depending on the system requirements and restrictions. Interestingly, since this change only implies changing the decision threshold, the same LLM can be run with different levels of protection using the proposed CLED. For example, when used to translate medical texts, we can allow for more than 20% re-computations to prioritize error detection than when used to translate news (at 1% overhead).

As an example, consider two different scenarios. In the first scenario, cost is the priority and only 1% overhead is allowed. In this case CLED can detect and correct 87% of the errors for T5 and 91% for OPUS-MT providing reasonable error coverage at low cost. In the second scenario, the objective is to achieve an error coverage of 99%, CLED requires 79% overhead for T5 and 74% for OPUS-MT. In this case, the saving versus recomputation that always detects all errors is not large and thus CLED is less attractive.

The overhead introduced by the concurrent error detection has not been included in the previous discussion. The rationale is that it is negligible. The results for the run times in Table 6 already show that the concurrent detector is 10x to 45x faster than the main model. The key observation is that language models are continuously increasing their complexity while the detector will remain the same. Therefore, the overheads, which are already small, will become negligible. To corroborate this observation, simple tasks have been run on a newer LLM, namely Mistral-7B. The execution time is 1.67 seconds, so the error detector is 150x times faster, more than in the smaller models considered in the evaluation.

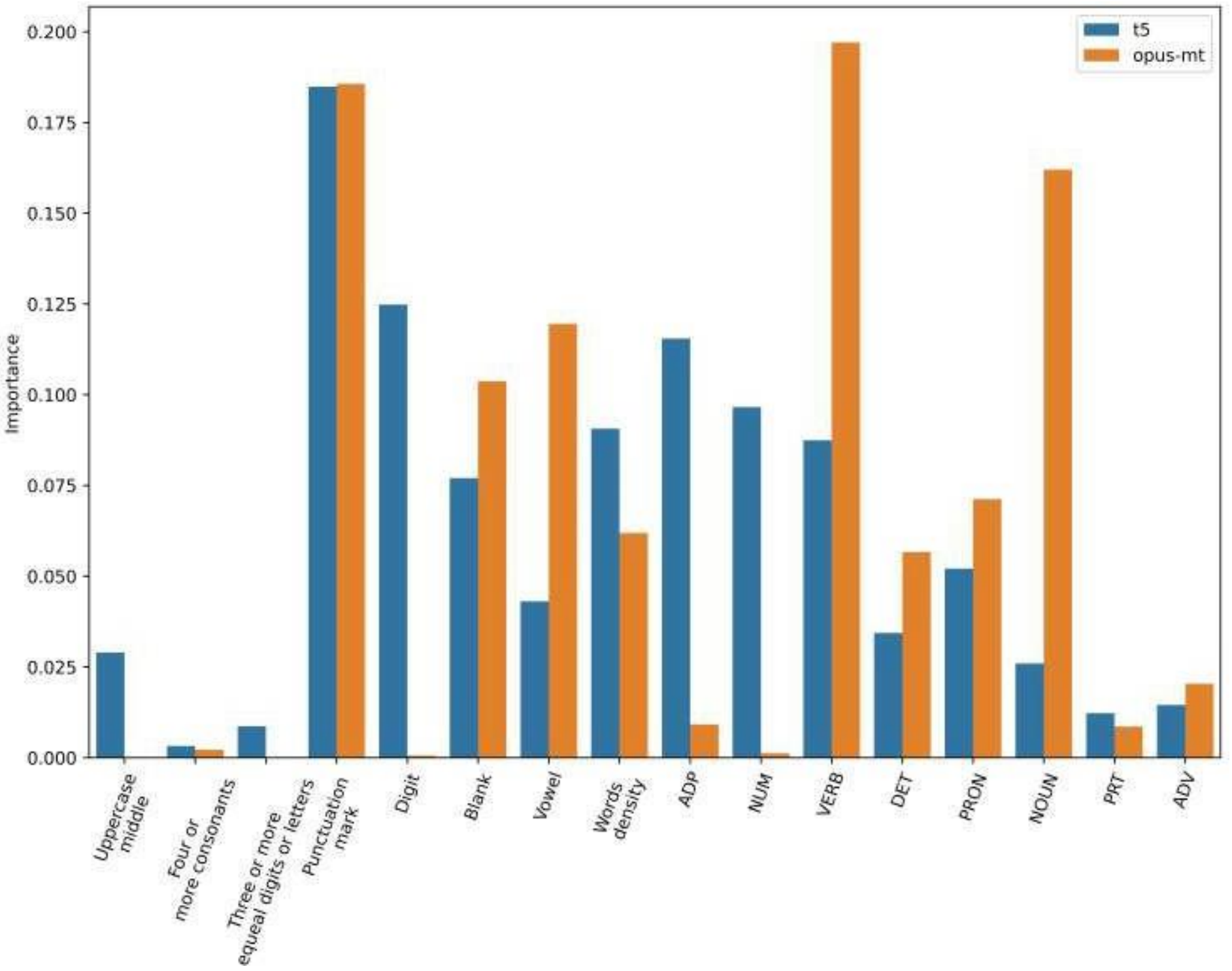

Fig. 11. Features importance for the Random Forest model in the CLED scheme.

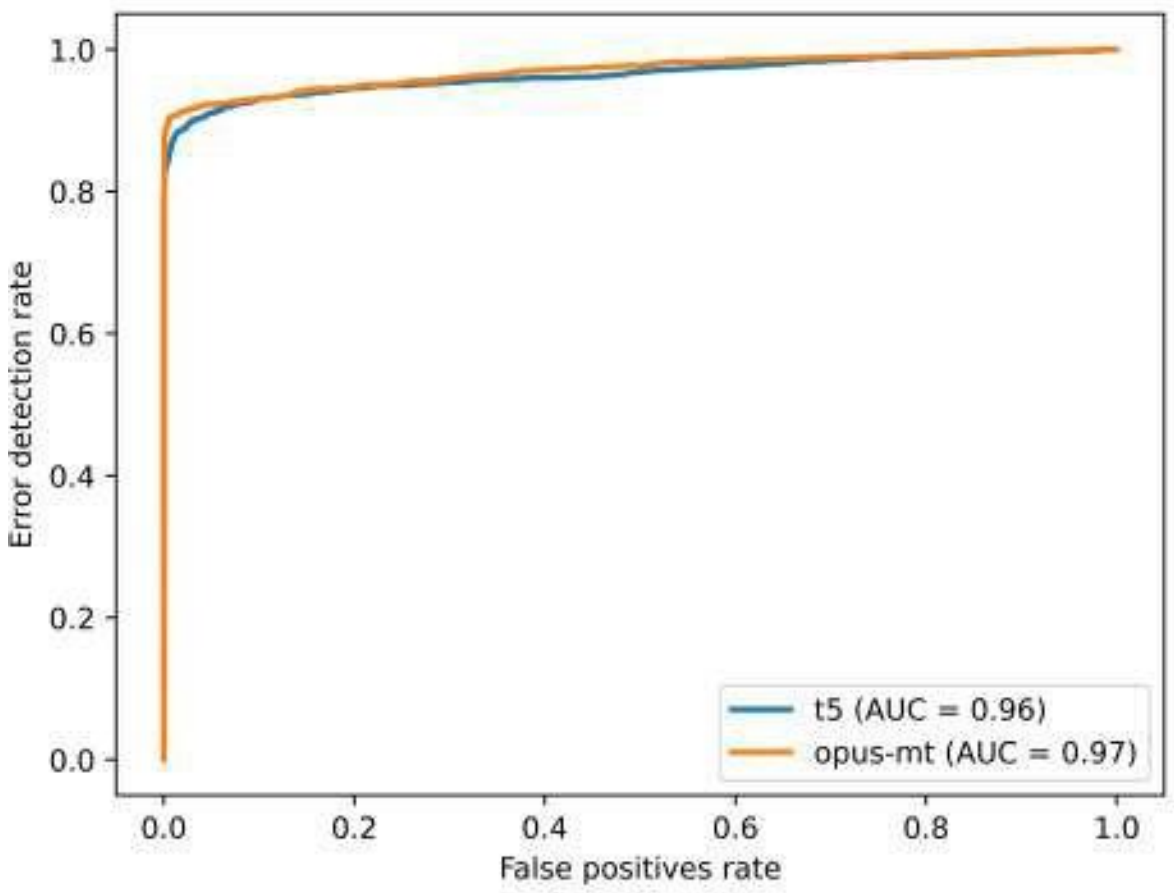

Fig. 12. Error detection rates for different percentages of re-computation (ROC curve).

After presenting all results and analysis, we can return to our initial research questions. For RQ1, the presented results show that linguistic features provide significant information on the presence of errors and can be used to efficiently detect errors using a simple classifier. Most errors can be detected and corrected with only 1% overhead. As for RQ2, the same features and classifiers have been used on two different LLMs and applications achieving similar performance levels. This suggests that linguistic error detection can be generally applicable to LLMs and easily integrated in their design flow.

## 6 Conclusion

This paper has presented a new CLED methodology to detect errors on LLMs at application level; it uses the rules of the language and the patterns of valid texts to detect texts generated by language models when they incur errors. The linguistic features are the inputs to a concurrent classifier that detects the errors. The proposed CLED-based scheme has been implemented and evaluated in two different models. The results show that more than 87% of the errors can be detected with only a 1% overhead. This shows the effectiveness of CLED that makes it attractive for low-cost error detection. More broadly, CLED can be applied to any LLM and its quantized versions making it a general scheme for LLM protection.

The CLED scheme can be refined and extended in different directions in future work. For example, it is of interest to evaluate the effectiveness of CLED for languages other than English and for additional tasks/models. In particular, the evaluation of CLED for widely used commercial models such as GPT4 or Gemini would be interesting, unfortunately, those models are not open source and thus error injection is not possible. Similarly, the evaluation of the effectiveness of CLED in multilingual and domain specific tasks is also of interest. Moreover, the investigation of additional linguistic features that can improve the detection rate is also of interest. Finally, a theoretical analysis study of the reasons behind the effectiveness of CLED is also of interest.

## APPENDIX

The abbreviations and acronyms used in the paper are listed in the following table.

TABLE 7
Abbreviations

| | |
|---|---|
| CA | Cross-attention |
| CB | Critical Bit |
| CED | Concurrent Error Detection |
| CCED | Concurrent Classifier Error Detection |
| FFT | Fast Fourier Transform |
| FFN | Feed-Forward Network |
| FPGA | Field Programmable Gate Array |
| GPU | Graphics Processing Units |
| LLM | Large Language Model |
| MH-SA | Multi-Head self-attention |
| NLP | Natural Language Processing |
| POS | Part-of-Speech |
| SA | Self-Attention |

## REFERENCES

[1] W. X. Zhao, K. Zhou, J. Li, T. Tang, X. Wang, Y. Hou, Y. Min, B. Zhang, J. Zhang, Z. Dong, Y. Du, C. Yang, Y. Chen, Z. Chen, J. Jiang, R. Ren, Y. Li, X. Tang, Z. Liu, P. Liu, J.-Y. Nie, J.-R. Wen, A survey of large language models, arXiv preprint 2303.18223 (2023). arXiv:2303.18223.

[2] J. Devlin, M.-W. Chang, K. Lee, K. Toutanova, BERT: Pre-training of Deep Bidirectional Transformers for Language Understanding (2019). arXiv:1810.04805.

[3] C. Raffel, N. Shazeer, A. Roberts, K. Lee, S. Narang, M. Matena, Y. Zhou, W. Li, P. J. Liu, Exploring the limits of transfer learning with a unified text-to-text transformer, Journal of Machine Learning Research 21 (140) (2020) 1–67.
URL http://jmlr.org/papers/v21/20-074.html

[4] OpenAI, Gpt-4 technical report, arXiv preprint 2310.06825 (2023). arXiv:2303.08774.

[5] G. Team, Gemini: A family of highly capable multimodal models, arXiv preprint 2312.11805 (2023). arXiv:2312.11805.

[6] H. T. et al, Llama 2: Open foundation and fine-tuned chat models, arXiv preprint 2307.09288 (2023). arXiv:2307.09288.

[7] A. Q. J. et al, Mistral 7b, arXiv preprint 2310.06825 (2023). arXiv:2310.06825.

[8] U. K. Agarwal, Resilience assessment of machine learning applications under hardware faults, Ph.D. thesis, University of British Columbia (2023).

[9] M. V. Beigi, S. Gurumurthi, V. Sridharan, Reliability, availability, and serviceability challenges for heterogeneous system design, in: 2022 IEEE International Reliability Physics Symposium (IRPS), 2022, pp. 2C.4–1–2C.4–8. doi:10.1109/IRPS48227.2022.9764554.

[10] S. S. Mukherjee, M. Kontz, S. K. Reinhardt, Detailed design and evaluation of redundant multithreading alternatives, ACM SIGARCH Computer Architecture News 30 (2) (2002) 99–110.

[11] M. A. Neggaz, I. Alouani, S. Niar, F. Kurdahi, Are cnns reliable enough for critical applications? an exploratory study, IEEE Design and Test 37 (2) (2020) 76–83. doi:10.1109/MDAT.2019.2952336.

[12] L. M. Luza, A. Ruospo, D. So¨ derstro¨ m, C. Cazzaniga, M. Kastriotou, E. Sanchez, A. Bosio, L. Dilillo, Emulating the effects of radiation-induced soft-errors for the reliability assessment of neural networks, IEEE Transactions on Emerging Topics in Computing 10 (4) (2022) 1867–1882. doi:10.1109/TETC.2021.3116999.

[13] C. Bolchini, L. Cassano, A. Miele, A. Toschi, Fast and accurate error simulation for cnns against soft errors, IEEE Transactions on Computers 72 (4) (2023) 984–997. doi:10.1109/TC.2022.3184274.

[14] Z. Gao, H. Zhang, Y. Yao, J. Xiao, S. Zeng, G. Ge, Y. Wang, A. Ullah, P. Reviriego, Soft error tolerant convolutional neural networks on fpgas with ensemble learning, IEEE Transactions on Very Large Scale Integration (VLSI) Systems 30 (3) (2022) 291–302. doi:10.1109/TVLSI.2021.3138491.

[15] Z. Gao, J. Wang, R. Su, P. Reviriego, S. Liu, L. Fabrizio, On the dependable operation of bidirectional encoder representations from transformers (bert) in the presence of soft errors, in: 2023 IEEE 23rd International Conference on Nanotechnology (NANO), 2023, pp. 582–586. doi:10.1109/NANO58406.2023.10231204.

[16] J. Tiedemann, S. Thottingal, OPUS-MT — Building open translation services for the World, in: Proceedings of the 22nd Annual Conference of the European Association for Machine Translation (EAMT), Lisbon, Portugal, 2020.

[17] L. Costas-Pe´rez, J. J. Rodr´ıguez-Andina, Algorithmic Concurrent Error Detection in Complex Digital-Processing Systems, IEEE Design and Test of Computers 26 (1) (2009) 60–67. doi:10.1109/MDT.2009.6.

[18] A. Mahmood, E. McCluskey, Concurrent error detection using watchdog processors-a survey, IEEE Transactions on Computers 37 (2) (1988) 160–174. doi:10.1109/12.2145.

[19] Z. Alkhalifa, V. Nair, N. Krishnamurthy, J. Abraham, Design and evaluation of system-level checks for on-line control flow error detection, IEEE Transactions on Parallel and Distributed Systems 10 (6) (1999) 627–641. doi:10.1109/71.774911.

[20] P. Reviriego, Z. Wang, A. Alonso, Z. Gao, F. Niknia, S. Liu, F. Lombardi, Concurrent classifier error detection (cced) in large scale machine learning systems, IEEE Transactions on Reliability (2024) 1–10doi:10.1109/TR.2024.3355408.

[21] X. Xue, C. Liu, Y. Wang, B. Yang, T. Luo, L. Zhang, H. Li, X. Li, Soft error reliability analysis of vision transformers, IEEE Transactions on Very Large Scale Integration (VLSI) Systems (2023).

[22] Z. Gao, L. Yuan, P. Reviriego, S. Liu, F. Lombardi, Dependability evaluation of stable diffusion with soft errors on the model parameters, arXiv preprint arXiv:2404.00352 (2024).

[23] U. K. Agarwal, A. Chan, K. Pattabiraman, Resilience assessment of large language models under transient hardware faults, in: 2023 IEEE 34th International Symposium on Software Reliability Engineering (ISSRE), IEEE, 2023, pp. 659–670.

[24] Z. Chen, G. Li, K. Pattabiraman, A low-cost fault corrector for deep neural networks through range restriction, in: 2021 51st Annual IEEE/IFIP International Conference on Dependable Systems and Networks (DSN), IEEE, 2021, pp. 1–13.

[25] Q. S. Sha, M. Paulitsch, K. Pattabiraman, K. Hagn, F. Oboril, C. Buerkle, K.-U. Scholl, G. Hinz, A. Knoll, Global clipper: Enhancing safety and reliability of transformer-based object detection models, arXiv preprint arXiv:2406.03229 (2024).

[26] A. Reddy, P. Banerjee, Algorithm-based fault detection for signal processing applications, IEEE Transactions on Computers 39 (10) (1990) 1304–1308. doi:10.1109/12.59860.

[27] R. Reiter, A theory of diagnosis from first principles, Artificial intelligence 32 (1) (1987) 57–95.

[28] M. Cheng, T. Sun, S. Nazarian, P. Bogdan, Trustworthiness evaluation and trust-aware design of cnn architectures, in: Conference on Lifelong Learning Agents, PMLR, 2022, pp. 1086–1102.

[29] K. Ito, Y. Zhang, H. Itsuji, T. Uezono, T. Toba, M. Hashimoto, Analyzing due errors on gpus with neutron irradiation test and fault injection to control flow, IEEE Transactions on Nuclear Science 68 (8) (2021) 1668–1674. doi:10.1109/TNS.2021.3098845.

[30] A. M. Keller, M. J. Wirthlin, Partial tmr for improving the soft error reliability of sram-based fpga designs, IEEE Transactions on Nuclear Science 68 (5) (2021) 1023–1031. doi:10.1109/TNS.2021.3070856.

[31] R. Baumann, Soft errors in advanced computer systems, IEEE Design and Test of Computers 22 (3) (2005) 258–266. doi:10.1109/MDT.2005.69.

[32] K. Blagec, G. Dorffner, M. Moradi, S. Ott, M. Samwald, A global analysis of metrics used for measuring performance in natural language processing, in: T. Shavrina, V. Mikhailov, V. Malykh, E. Artemova, O. Serikov, V. Protasov (Eds.), Proceedings of NLP Power! The First Workshop on Efficient Benchmarking in NLP, Association for Computational Linguistics, Dublin, Ireland, 2022, pp. 52–63. doi:10.18653/v1/2022.nlppower-1.6.
URL https://aclanthology.org/2022.nlppower-1.6/

[33] M. Cettolo, M. Federico, L. Bentivogli, J. Niehues, S. Stu¨ker, K. Sudoh, K. Yoshino, C. Federmann, Overview of the IWSLT 2017 evaluation campaign, in: Proceedings of the 14th International Conference on Spoken Language Translation, International Workshop on Spoken Language Translation, Tokyo, Japan, 2017, pp. 2–14.
URL https://aclanthology.org/2017.iwslt-1.1

[34] X. Xiao, C. Zhou, H. Ping, D. Cao, Y. Li, Y. Zhou, S. Li, P. Bogdan, Exploring neuron interactions and emergence in llms: From the multifractal analysis perspective, arXiv e-prints (2024) arXiv–2402.
[35] G. Brooks, Dictionary of the British English Spelling System, 1st Edition, Open Book Publishers, 2015.
[36] J. H. Clegg, W. C. Fails, Manual de fone´tica y fonolog´ıa espan˜olas, Routledge (2018) 111.
[37] M. G. O'Brien, S. M. Fagan, German phonetics and phonology: Theory and practice, Yale University Press (2016) 65.
[38] E. van Lier, The Oxford Handbook of Word Classes, Oxford University Press, 2023. doi:10.1093/oxfordhb/9780198852889.001.0001.
URL https://doi.org/10.1093/oxfordhb/9780198852889.001.0001
[39] N. Agarwal, M. A. Wani, P. Bours, Lex-pos feature-based grammar error detection system for the english language, Electronics 9 (10) (2020). doi:10.3390/electronics9101686.
URL https://www.mdpi.com/2079-9292/9/10/1686
[40] A. Mendhakar, D. H. S, Parts-of-speech (pos) analysis and classification of various text genres, Corpus-based Studies across Humanities (2023). doi:doi:10.1515/csh-2023-0002.
URL https://doi.org/10.1515/csh-2023-0002
[41] N. Chomsky, Three models for the description of language, IRE Transactions on information theory 2 (3) (1956) 113–124.

## BIOGRAPHIES

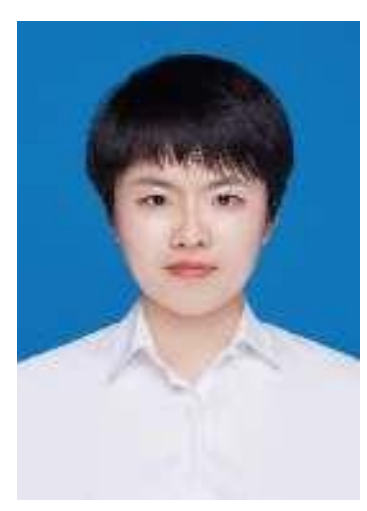

**Jinhua Zhu** received the PhD degree in Electrical and Information Engineering from Tianjin University, Tianjin, China, in 2024. She is currently with the School of Integrated Circuit Science and Engineering, Tianjin University of Technology, Tianjin, China. Her research focus now is fault-tolerant design for digital signal processing, data sketching and machine learning.

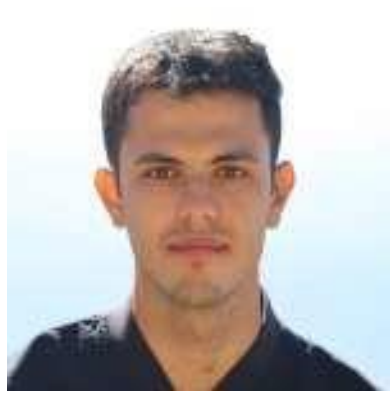

**Javier Conde** is an assistant professor at the Universidad Polite´cnica de Madrid, Madrid, Spain. His research interests include Digital Twins, Big Data, Semantic Web, and Artificial Intelligence.

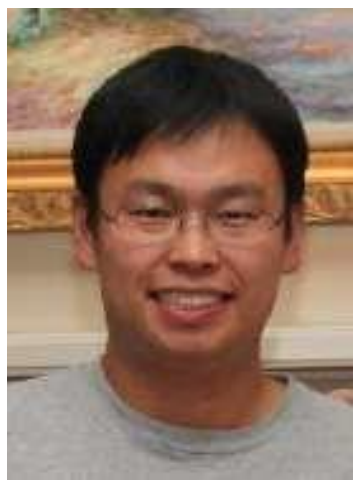

**Zhen Gao** received the BS, MS and PhD degree in Electrical and Information Engineering from Tianjin University, China, in 2005, 2007 and 2011, respectively. He was a visiting scholar in GeorgiaTech during 2008.10 2010.11, and a Postdoc researcher in the in Tsinghua University, China, during 2011.7 2014.11. Since 2014.12, he is an Associate Professor in Tianjin University. His focus now includes fault-tolerant DSPs design and Blockchain technologies.

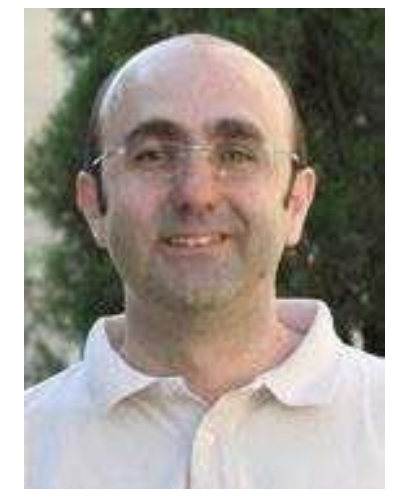

**Pedro Reviriego** received the M.Sc. and Ph.D. degrees in Telecommunications engineering from the Technical University of Madrid, Madrid, Spain, in 1994 and 1997, respectively. From 1997 to 2000, he was an R&D Engineer with Teldat, Madrid, working on router implementation. In 2000, he joined Massana to work on the development of 1000BaseT transceivers. From 2004 to 2007, he was a Distinguished Member of Technical Staff with the LSI Corporation, working on the development of Ethernet transceivers. From 2007 to 2018 he was with Universidad Antonio de Nebrija, and from 2018 to 2022 with Universidad Carlos III de Madrid. He is currently with Universidad Polite´cnica de Madrid.

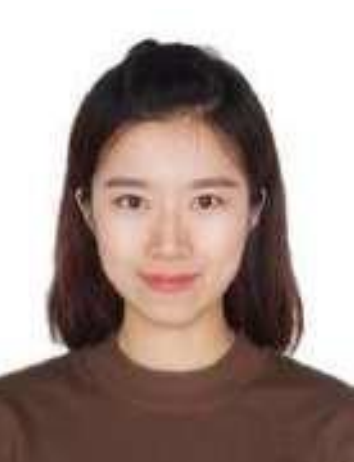

**Shanshan Liu** received the Ph.D. degree in Microelectronics and Solid-State Electronics from Harbin Institute of Technology, Harbin, China, in 2018. She was a post-doctoral researcher with the Department of Electrical and Computer Engineering (ECE), Northeastern University, Boston, USA, from 2018 to 2021, and an assistant professor with the Klipsch School of ECE, New Mexico State University, Las Cruces, USA from 2021 to 2023. She is currently with the School of Information and Communication Engineering, University of Electronic Science and Technology of China, Chengdu, China. Her research interests include fault tolerance design in high performance computing systems, emerging computing, VLSI design, dependable machine learning, error correction codes.

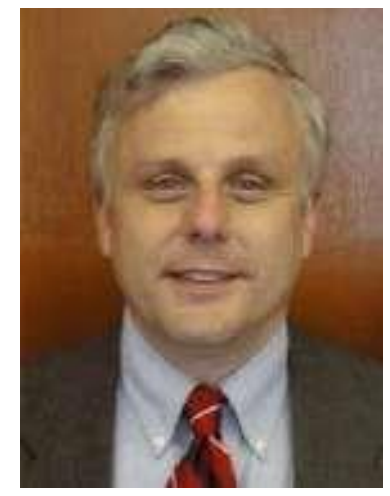

**Fabrizio Lombardi** received the B.Sc. degree (Hons.) in electronic engineering from the University of Essex, U.K., in 1977, the master's degree in microwaves and modern optics, University College London, in 1978, and the Ph.D. degree from the University of London in 1982. He is currently the International Test Conference (ITC) Endowed Chair Professorship with Northeastern University, Boston, USA. His research interests are bio-inspired and nano manufacturing/computing, VLSI design, testing, and fault/defect tolerance of digital systems. He is currently the 2022/23 President of the IEEE Nanotechnology Council and a member of the IEEE Publication Services and Products Board. He is a life fellow of the IEEE.